\documentclass[preprint,5p]{elsarticle}

\usepackage{natbib}
\usepackage{graphicx}          
\usepackage{xcolor}

\UseRawInputEncoding
\usepackage{graphicx}      
\usepackage[maxfloats=256]{morefloats}
\usepackage{hyperref}
\hypersetup{
    colorlinks=true,
    linkcolor=blue,
    filecolor=magenta,      
    urlcolor=cyan,
}
\newlength{\figureheight}
\newlength{\figurewidth}
\graphicspath{{./}{figure/}}
\usepackage{color}
\usepackage{amsmath} 
\usepackage{amssymb}  
\usepackage{bm}  
\usepackage{makecell}
\usepackage{array}
\usepackage{arydshln}
\usepackage{nicematrix}
\usepackage{siunitx}

\DeclareMathOperator*{\argmin}{arg\,min}
\usepackage{xspace}

\providecommand{\kuka}{\textsc{KUKA} LBR iiwa R820\xspace}
\usepackage{algorithm}
\usepackage{algpseudocode}
\usepackage{mathrsfs}
\usepackage{etoolbox}
\makeatletter
\def\BState{\State\hskip-\ALG@thistlm}
\makeatother
\usepackage{tikz}
\usepackage{adjustbox}
\usepackage{pgfplots}
\usepgfplotslibrary{groupplots}
\usetikzlibrary{pgfplots.groupplots}
\usetikzlibrary{plotmarks}
\usetikzlibrary{calc}
\usepgfplotslibrary{external}
\pgfplotsset{compat=newest}

\usetikzlibrary{matrix}
\pgfplotsset{grid style={dashed,gray}}
\usepackage{svg-extract}
\usepackage{mathtools}
\usepackage{cleveref}

\DeclareMathOperator{\arctantwo}{arctan2}

\allowdisplaybreaks[1]

\journal{Mechatronics}

\begin{document}
\begin{frontmatter}

\title{Machine Learning-based Framework for Optimally Solving the Analytical Inverse Kinematics for Redundant Manipulators}

\author[tuwien]{M.N. Vu\corref{cor1}}\ead{vu@acin.tuwien.ac.at}  
\author[tuwien]{F. Beck}\ead{beck@acin.tuwien.ac.at}
\author[tuwien]{M. Schwegel}\ead{schwegel@acin.tuwien.ac.at}
\author[tuwien]{C. Hartl-Nesic}\ead{hartl@acin.tuwien.ac.at}
\author[liverpool]{A. Nguyen}\ead{anh.nguyen@liverpool.ac.uk}
\author[tuwien,ait]{A. Kugi}\ead{kugi@acin.tuwien.ac.at}              
\cortext[cor1]{Corresponding author}
\address[tuwien]{Automation \& Control Institute (ACIN), TU Wien, Vienna, Austria}  
\address[liverpool]{Department of Computer Science, University of Liverpool, Liverpool, England}
\address[ait]{Center for Vision, Automation \& Control, Austrian Institute of Technology GmbH (AIT), Vienna, Austria}


\begin{abstract}                          
Solving the analytical inverse kinematics (IK) of redundant manipulators in real time is a difficult problem in robotics since its solution for a given target pose is not unique. 
Moreover, choosing the optimal IK solution with respect to application-specific demands helps to improve the robustness and to increase the success rate when driving the manipulator from its current configuration towards a desired pose. 
This is necessary, especially in high-dynamic tasks like catching objects in mid-flights.
To compute a suitable target configuration in the joint space for a given target pose in the trajectory planning context, various factors such as travel time or manipulability must be considered. However, these factors increase the complexity of the overall problem which impedes real-time implementation. 
In this paper, a real-time framework to compute the analytical inverse kinematics of a redundant robot is presented. 
To this end, the analytical IK of the redundant manipulator is parameterized by so-called redundancy parameters, which are combined with a target pose to yield a unique IK solution. 
Most existing works in the literature either try to approximate the direct mapping from the desired pose of the manipulator to the solution of the IK or cluster the entire workspace to find IK solutions. 
In contrast, the proposed framework directly learns these redundancy parameters by using a neural network (NN) that provides the optimal IK solution with respect to the manipulability and the closeness to the current robot configuration. Monte Carlo simulations show the effectiveness of the proposed approach which is accurate and real-time capable ($\approx$ \SI{32}{\micro\second}) on the KUKA  LBR iiwa 14 R820. \end{abstract}
\end{frontmatter}
\section{Introduction}
The inverse kinematics (IK) \cite{spong2005robot} solution is fundamental for many applications in robotics involving motion planning, e.g., point-to-point trajectory optimization \cite{wang2019smooth, KRAMER2021102523}, path-wise trajectory planning \cite{9616379,GATTRINGER2022102753}, dexterous grasping \cite{han2020local,qiu2021precision}, and pick-and-place scenarios \cite{saut2010planning,PELLICCIARI2013326}. Solving the IK problem for a given target position in the task space yields the robot's configuration in joint space that satisfies the kinematic constraints \cite{lynch2017modern}.

There are three types of techniques to solve IK problems, i.e. the algebraic approach, see, e.g., \cite{raghavan1993inverse, cox2013ideals}, the analytical (or so-called geometric) approach, see, e.g., \cite{shimizu2008analytical,wiedmeyer2020real}, and the numerical (or so-called iterative) approach, see, e.g., \cite{safeea2021modified, siciliano1990kinematic}. In the algebraic approach, essentially systems of polynomial equations \cite{paul1979kinematic} are solved. Typically, they are classified as difficult algebraic computational problems \cite{cox2013ideals}. In general, this algebraic problem can be solved for a manipulator with 6 degrees of freedom (DoF), see, e.g., \cite{diankov2008openrave}, \textcolor{black}{but is not generally applicable to kinematically redundant manipulators \cite{sciavicco1988solution}.} 
On the other hand, numerical methods, typically based on least-squares or pseudoinverse formulations, are widely employed, see, e.g., \cite{XING2021102639}, for various kinematic structures due to their simplicity, low computing time, and their generality. However, these methods may converge to a local minimum that predominantly depends on the initial guess of the solution. 

In contrast to the numerical IK, the analytical IK computes the exact solution, which is important for many industrial applications \cite{zhou2014mobile,ghafil2018research}. The computing time of the analytical IK solutions is much faster and real-time capable, compared to the numerical approach. 
While the analytical IK is only available for specific robot kinematics, most industrial robots are designed such that an analytical solution of the IK is available. Hence, the IK of 6-DoF industrial robots with a spherical wrist, non-offset $7$ DoF S-R-S redundant manipulators \cite{taki2014novel,kuhlemann2016robust}, e.g., KUKA LBR iiwa 14 R820, but also the offset redundant manipulator, e.g., Franka Emika Panda \cite{franka_panda}, OB7 \cite{OB7}, and Neura Robotics LARA \cite{LARA}, can be solved analytically. These manipulators are often referred to as collaborative robots (Cobot). Typically, the analytical IK parameterizes the robot redundancy by additional (three) parameters, which are usually named redundancy parameters. Examples are \cite{shimizu2008analytical} and \cite{he2021analytical} for the non-offset and offset redundant manipulator, respectively.
Due to the redundant nature of these parameters, infinite sets of parameters exist in general yielding different IK solutions. However, finding the set of redundancy parameters which represents the $\emph{best}$ IK solution of a specific task is a non-trivial problem.

In this work, a learning-based framework to compute the optimal set of redundancy parameters for an analytical IK is proposed. 
This improves the computing performance for solving the IK problem, which is essential for highly dynamic tasks like catching objects in mid-flight or handing over objects between moving agents. These tasks are frequently solved using trajectory optimization where typically dynamic system constraints, state and control input constraints, and a target pose constraint are considered. 

The target pose constraint is often formulated in the task space and for kinematically redundant robots, in which infinite joint configurations satisfy this constraint. Utilizing the analytical IK in the trajectory optimization problem allows to find the best target joint configuration for a specific task. On the other hand, computing time becomes an issue with this approach.
Recently, \cite{wiedmeyer2020real} proposed a real-time capable closed-form solution for the KUKA iiwa 14 R820, where the authors minimize the joint velocities and accelerations while avoiding joint boundaries for a trajectory tracking task. This approach does neither consider the dynamic constraints nor the system state and input constraints in the trajectory optimization. 

\begin{figure}
    \centering
    \def\svgwidth{1\columnwidth}
\begingroup%
  \makeatletter%
  \providecommand\color[2][]{%
    \errmessage{(Inkscape) Color is used for the text in Inkscape, but the package 'color.sty' is not loaded}%
    \renewcommand\color[2][]{}%
  }%
  \providecommand\transparent[1]{%
    \errmessage{(Inkscape) Transparency is used (non-zero) for the text in Inkscape, but the package 'transparent.sty' is not loaded}%
    \renewcommand\transparent[1]{}%
  }%
  \providecommand\rotatebox[2]{#2}%
  \newcommand*\fsize{\dimexpr\f@size pt\relax}%
  \newcommand*\lineheight[1]{\fontsize{\fsize}{#1\fsize}\selectfont}%
  \ifx\svgwidth\undefined%
    \setlength{\unitlength}{994.97793123bp}%
    \ifx\svgscale\undefined%
      \relax%
    \else%
      \setlength{\unitlength}{\unitlength * \real{\svgscale}}%
    \fi%
  \else%
    \setlength{\unitlength}{\svgwidth}%
  \fi%
  \global\let\svgwidth\undefined%
  \global\let\svgscale\undefined%
  \makeatother%
  \begin{picture}(1,0.68623537)%
    \lineheight{1}%
    \setlength\tabcolsep{0pt}%
    \put(0,0){\includegraphics[width=\unitlength,page=1]{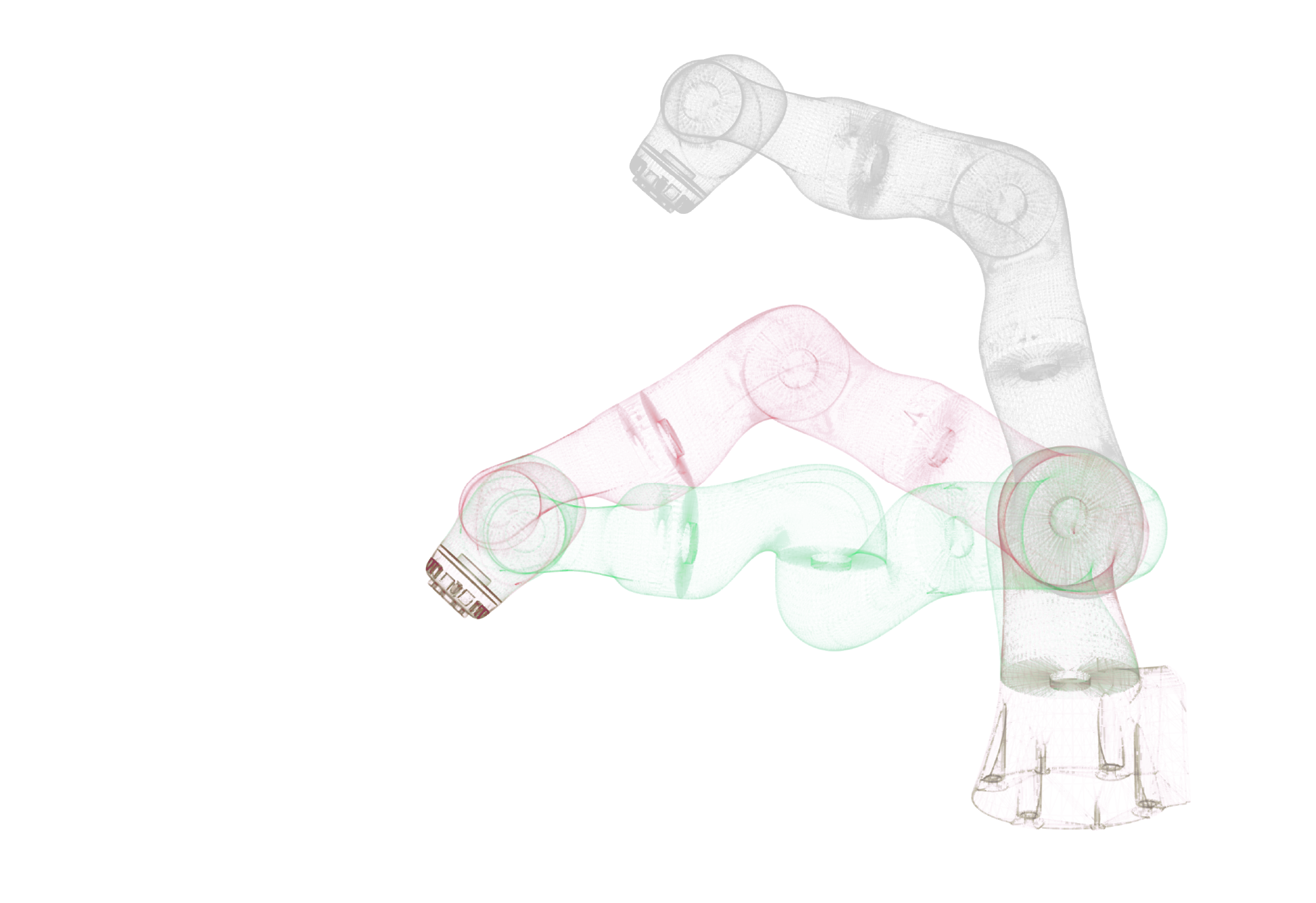}}%
    \put(0.00221843,0.63514668){\color[rgb]{0,0,0}\makebox(0,0)[lt]{\lineheight{1.25}\smash{\begin{tabular}[t]{l}initial configuration\end{tabular}}}}%
    \put(0,0){\includegraphics[width=\unitlength,page=2]{iiwa_overview_application.pdf}}%
    \put(0.06099069,0.34706025){\color[rgb]{0,0,0}\makebox(0,0)[lt]{\lineheight{1.25}\smash{\begin{tabular}[t]{l}target pose \end{tabular}}}}%
    \put(0,0){\includegraphics[width=\unitlength,page=3]{iiwa_overview_application.pdf}}%
    \put(-0.00201954,0.50336182){\color[rgb]{0,0,0}\makebox(0,0)[lt]{\lineheight{1.25}\smash{\begin{tabular}[t]{l}possible target \\configurations\end{tabular}}}}%
    \put(0,0){\includegraphics[width=\unitlength,page=4]{iiwa_overview_application.pdf}}%
  \end{picture}%
\endgroup%

    \caption{An example of a handover task between robot and human.}%
    \label{fig: example handover task}%
\end{figure}

In addition, it is crucial that the target configuration is well chosen among the infinite solutions of the IK, i.e. close to the initial robot configuration and with high manipulability. For example in a dynamic handover of an object where the target is moving, see Fig. \ref{fig: example handover task}, it is advantageous to choose a target configuration that maximizes manipulability such that the robot end-effector can move to another target configuration with high agility. 
Moreover, choosing a target configuration that is close to the initial configuration of the robot can lead to high performance of the trajectory optimization with a high success rate. Including such criteria in the trajectory optimization is the motivation for the work in this paper. 

To this end, a learning-based framework to include additional, application-specific criteria in the analytical IK of redundant robots is proposed. First, a database of $10^8$ random pairs of initial configurations and target poses is generated. For each pair, the optimal trajectory is computed using classical approaches, considering application-specific criteria, and is stored in the database together with the set of optimal redundancy parameters. This database serves as the basis to train a neural network (NN), which is used to predict the optimal redundancy parameters for the analytical IK in highly dynamic real-time applications.



The main contribution of this work is a learning-based framework that employs a NN to predict the redundancy parameters of an analytical IK. This yields an optimal target joint configuration for a given target pose by considering application-specific criteria. In this work, the target joint configuration is chosen close to the current joint configuration of the robot and to have a high measure of manipulability. The proposed learning framework significantly speeds up the computing time of the trajectory optimization problem.    
Note that the proposed framework is tailored to the non-offset redundant manipulator KUKA LBR iiwa 14 R820. Nevertheless, it is also applicable to other kinematically redundant robots with an analytical IK, e.g., \cite{franka_panda,OB7,LARA}. 

The remainder of this paper is organized as follows.
Section \ref{section: Trajectory Optimization} presents the mathematical modeling and analytical
inverse kinematics. Additionally, details of the point-to-point (PTP) trajectory optimization problem and the algorithm for determining the optimal target joint configuration w.r.t. application-specific criteria are given. The learning framework for predicting the redundancy parameters for the analytical IK problem
including database generation and the proposed NN is presented in Section \ref{section: Framework for learning analytic inverse kinematics parameters}. Simulation results
are shown in Section \ref{section: simulation results}. The last section, Section \ref{section: conclusion}, concludes this work.

\section{Trajectory Optimization Framework}
\label{section: Trajectory Optimization}
This section presents the trajectory optimization framework which is commonly used in robotics \cite{betts1998survey}. For example, in Figure \ref{fig: example handover task}, with a given target pose and a robot's initial configuration, an optimal trajectory in joint space is planned for the robot to catch an object. 

In this section, the mathematical modeling of the KUKA LBR iiwa 14 R820, including kinematics and system dynamics, is briefly summarized. Then, the analytical inverse kinematics with the redundancy parameters of this redundant manipulator is presented in Section \ref{subsection: Analytic inverse kinematics}. Subsequently, a point-to-point trajectory optimization is performed, which is used to plan trajectories to the optimal target configuration, explained in Section 
\ref{subsection: optimal target conf}. 
\subsection{Mathematical modeling}
\label{subsection: Mathematical modeling}
\begin{figure}
    \centering
    \def\svgwidth{0.5\columnwidth}
    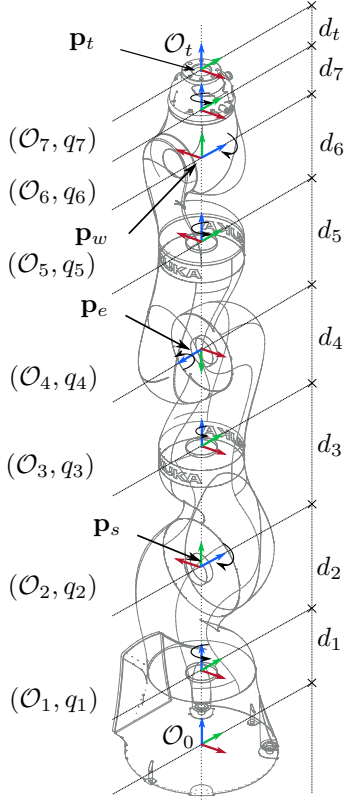
    \caption{Schematic drawing of the robot KUKA LBR iiwa. The $x$-, $y$-, and $z$-axis of each coordinate frame are shown as red, green, and blue arrows, respectively.}%
    \label{fig: coordinate}%
\end{figure}

The KUKA LBR iiwa 14 R820 is an anthropomorphic manipulator due to its similarity to a human arm, which has an S-R-S kinematic structure \cite{taki2014novel}. The coordinate frames $\mathcal{O}_i$ and the corresponding seven revolute joints $q_i, \: i=1,\ldots,7,$ of the robot are shown in Fig. \ref{fig: coordinate}. The red, green, and blue arrows represent the $x$-, $y$-, and $z$-axis, respectively. 
The shoulder intersection position $\mathbf{p}_s$ of the joint axes $q_1$, $q_2$, and $q_3$ and the wrist intersection position $\mathbf{p}_w$ of the joint axes $q_5$, $q_6$, and $q_7$ correspond to the shoulder and wrist positions of the human arm, respectively. The elbow position $\mathbf{p}_e$ is in the center of the joint axis $q_4$. 

\begin{table}
\caption{Coordinate transformation of the robot}
\label{table: coordinate transformation}
\begin{center}
\scalebox{1}{
\begin{tabular}{c c c}
\makecell{\textbf{Frame} \\$\mathcal{O}_n$} &\makecell{ \textbf{Frame}\\ $\mathcal{O}_m$ }& \makecell{\textbf{Transformation matrix} \\$\mathbf{T}_{n}^{m}$}\\ 
\hline
0 & 1 & $\mathbf{T}_{\mathbf{D},z}(d_1) \mathbf{T}_{\mathbf{R},z}(q_1)$  \\  
1 & 2 & $\mathbf{T}_{\mathbf{D},z}(d_2)\mathbf{T}_{\mathbf{R},z}(-\pi)\mathbf{T}_{\mathbf{R},x}(\pi/2)\mathbf{T}_{\mathbf{R},z}(q_2)$  \\  
2 & 3 & $\mathbf{T}_{\mathbf{D},y}(d_3)\mathbf{T}_{\mathbf{R},z}(\pi)\mathbf{T}_{\mathbf{R},x}(\pi/2)\mathbf{R}_z(q_3)$  \\  
3 & 4 & $\mathbf{T}_{\mathbf{T},z}(d_4) \mathbf{T}_{\mathbf{R},x}(\pi/2)\mathbf{T}_{\mathbf{R},z}(q_4)$ \\  
4 & 5 & $\mathbf{T}_{\mathbf{D},y}(d_5) \mathbf{T}_{\mathbf{R},z}(\pi)\mathbf{T}_{\mathbf{R},x}(\pi/2)\mathbf{T}_{\mathbf{R},z}(q_5)$  \\  
5 & 6 & $\mathbf{T}_{\mathbf{D},y}(d_6) \mathbf{T}_{\mathbf{R},x}(\pi/2)\mathbf{T}_{\mathbf{R},z}(q_6)$  \\  
6 & 7 & $\mathbf{T}_{\mathbf{D},z}(d_7) \mathbf{T}_{\mathbf{R},z}(\pi)\mathbf{T}_{\mathbf{R},x}(\pi/2)\mathbf{T}_{\mathbf{R},z}(q_7)$  \\  
7 & t & $\mathbf{T}_{\mathbf{D},z}(d_t)$  \\  
\hline
\end{tabular}
}
\end{center}
\end{table}

\begin{table}[h]
\caption{Kinematic and dynamic limits of the system}
\label{table: constraints}
\begin{center}
\scalebox{1}{
\begin{tabular}{c c c c c }
\textbf{Joint} $i$ & \textbf{Joint limits} & \textbf{Velocity limits} & \textbf{Torque limits} \\
 & $\overline{q_i}$ ($^{\circ}$) & $\overline{\dot{q}_i}$ ($^{\circ}$/\SI{}{\second}) & $\overline{\tau_{i}}$ (\SI{}{\newton}) \\
\hline
1 & 170 & 85 & 320 \\
2 & 120 & 85 & 320 \\
3 & 170 & 100 & 176 \\
4 & 120 & 75 & 176 \\
5 & 170 & 130 & 110 \\
6 & 120 & 135 & 40 \\
7 & 175 & 135 & 40 \\
\hline
\end{tabular}
}
\end{center}
\end{table}

The robot is modeled as a rigid-body system with the generalized coordinates $\mathbf{q}^\mathrm{T} = [q_1,q_2,\dots,q_7]$, see Fig. \ref{fig: coordinate}, which are the rotation angles $q_i$ around the $z$-axes (blue arrows) of each coordinate frame $\mathcal{O}_i$, $i=1,\dots,7$. 
To describe the kinematic relationship between the joint angles and the pose of the robot links comprising position and orientation, the homogeneous transformations $\mathbf{T}_n^m$ between two adjacent frames $\mathcal{O}_n$ and $\mathcal{O}_m$ are constructed, see Tab. \ref{table: coordinate transformation}.
Here and in the following, the homogeneous transformation of a simple translation by distance $d$ along the local axis $j \in \{x, y,z\}$ is denoted by $\mathbf{T}_{\mathbf{D},j}(d)$, while an elementary rotation around the local axis $j \in \{x,y,z\}$ by the angle $\phi$ is described by $\mathbf{T}_{\mathbf{R},j}(\phi)$. 
The end-effector transformation matrix $\mathbf{T}_0^e$ is referred to as forward kinematics and is computed in the form
\begin{equation}
\mathbf{T}_0^e = \mathrm{FK}(\mathbf{q}) = \prod_{i=0}^{7}\mathbf{T}_{i}^{i+1} = \begin{bmatrix}
\mathbf{R}_0^7(\mathbf{q}) & \mathbf{p}_t(\mathbf{q}) \\
\mathbf{0} & \mathbf{1}
\end{bmatrix} 
\label{eq: FK}
\end{equation}
comprising the $3$D tip position $\mathbf{p}_t \in \mathbb{R}^3$ and the $3$D orientation of the end effector as rotation matrix $\mathbf{R}_0^7 \in \mathbb{R}^{3\times3}$. 
The equations of motion are derived using the Lagrange formalism, see, e.g., \cite{spong2005robot}, 
\begin{equation}
\mathbf{M}(\mathbf{q})\ddot{\mathbf{q}} + \mathbf{C}(\mathbf{q},\dot{\mathbf{q}})\dot{\mathbf{q}} + \mathbf{g}(\mathbf{q}) = 
\bm{\tau},
\label{eq: EOM}
\end{equation}
where $\mathbf{M}(\mathbf{q})$ denotes the symmetric and positive definite mass matrix, $\mathbf{C}(\mathbf{q}, \dot{\mathbf{q}})$ is the Coriolis matrix, $\mathbf{g}(\mathbf{q})$ is the force vector associated with the potential energy, and $\bm{\tau}$ are the motor torque inputs. The kinematic and dynamic parameters of the KUKA LBR iiwa in (\ref{eq: EOM}) are taken from \cite{sturz2017parameter}. Since the mass matrix $\mathbf{M}(\mathbf{q})$ is invertible, (\ref{eq: EOM}) is rewritten in the state-space form 
\begin{equation}
\dot{\mathbf{x}} = 
\begin{bmatrix}
\dot{\mathbf{q}} \\
\mathbf{M}^{-1}(\mathbf{q}) (\bm{\tau}-\mathbf{C}(\mathbf{q},\dot{\mathbf{q}})\dot{\mathbf{q}}-\mathbf{g}(\mathbf{q}))
\end{bmatrix},
\label{eq: state space}
\end{equation} 
with the system state $\mathbf{x}^{\mathrm{T}} = [\mathbf{q}^{\mathrm{T}},\dot{\mathbf{q}}^{\mathrm{T}}]$. 
The kinematic and dynamic limits of the robot \cite{kukalbriiwa} are summarized in Tab. \ref{table: constraints}. All limits are symmetric w.r.t. zero, i.e. $\underline{q_i} = -\overline{q_i}$, $\underline{\dot{q}_i} = -\overline{\dot{q}_i}$, and $\underline{\tau_{i}} = -\overline{\tau_{i}}$.

To reduce the complexity of the system dynamics (\ref{eq: EOM}), the vector of joint acceleration $\ddot{\mathbf{q}}$ is utilized as a new control input for planning a trajectory in Section \ref{subsection: Point-to-point trajectory optimization}, i.e., $\mathbf{u}$ = $\ddot{\mathbf{q}} = \mathbf{M}^{-1}(\mathbf{q}) (\bm{\tau}-\mathbf{C}(\mathbf{q}, \dot{\mathbf{q}})\dot{\mathbf{q}}-\mathbf{g}(\mathbf{q}))$. Hence, the system dynamics (\ref{eq: state space}) is rewritten in the compact form
\begin{equation}
\dot{\mathbf{x}} = 
\begin{bmatrix}
\dot{\mathbf{q}} \\
\mathbf{u}
\end{bmatrix}.
\label{eq: state space 2}
\end{equation} 
\subsection{Analytical inverse kinematics}
\label{subsection: Analytic inverse kinematics}

Typically in manipulation tasks, the desired end-effector pose for a point-to-point motion is given in the $6$D Cartesian space in the form, cf. (\ref{eq: FK})
 \begin{equation}
\mathbf{T}_{0,d}^e = \begin{bmatrix}
\mathbf{R}_{0,d}^e & \mathbf{p}_{t,d} \\
\mathbf{0} & 1
\end{bmatrix}.
\label{eq: T_0_e}
\end{equation}
To compute the robot joint configuration $\mathbf{q}$  from a desired end-effector pose $\mathbf{T}_{0,d}^e$, the inverse kinematics (IK) of the robot has to be solved. In the following, an inverse kinematics solution with redundancy parmeters tailored to the non-offset 7-DoF robot KUKA LBR iiwa 14 R820 is shortly revisited, see, e.g., \cite{shimizu2008analytical}. Similar to \cite{kuhlemann2016robust,faria2018position}, the redundancy parameters of this robot are chosen as the binary vector $\mathbf{j}_c^\mathrm{T} = [j_s,j_e,j_w]$ and the angle $\varphi$, which are introduced below.

\begin{figure*}
    \centering
    \def\svgwidth{1.2\columnwidth}
    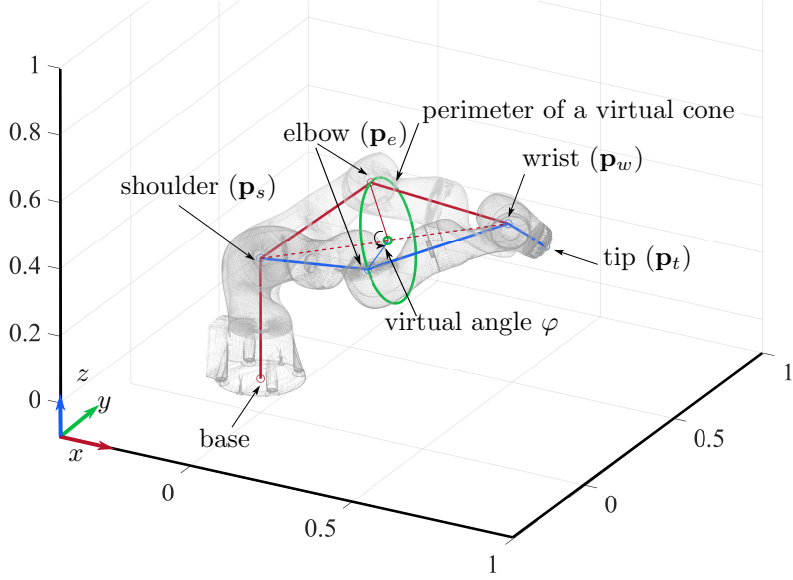
    \caption{Two configurations of the KUKA iiwa at the same pose are illustrated in red and blue lines. The green rim indicates the virtual movement of the elbow position w.r.t the specific end-effector pose. The red lines and blue lines illustrate the robot at the arm angle $\varphi = 0$ and  $\varphi = 95^{\circ}$, respectively.}%
    \label{fig: IK visual}%
\end{figure*}

With a given end-effector pose $\mathbf{T}_{0,d}^e$, the position of the robot wrist $\mathbf{p}_w$ in the world frame is fixed and is computed as
\begin{equation}
\mathbf{p}_{w} = \mathbf{p}_{t,d} - \mathbf{R}_{0,d}^7\begin{bmatrix}
0 & 0 & d_7+d_t
\end{bmatrix}^\mathrm{T},
\label{eq: p_w}
\end{equation}
with the distance from the wrist point to the end effector of the robot $d_7+d_t$. The vector $\mathbf{p}_{sw}$ from the fixed shoulder position $\mathbf{p}_s = \begin{bmatrix}
0 & 0 & d_1 + d_2
\end{bmatrix}^\mathrm{T}$ to the wrist position $\mathbf{p}_{w}$ is expressed as $\mathbf{p}_{sw} = \mathbf{p}_w - \mathbf{p}_s$. 
Using the law of cosines in the triangle formed by the shoulder, elbow, and wrist, the joint position $q_4$ is immediately calculated as 
\begin{equation}
 q_4 = j_e \arccos\left(\dfrac{\lvert\mathbf{p}_{sw}\rvert^2 - d_{se}^2 - d_{ew}^2}{2d_{se}d_{ew}}\right),
\label{eq: q_4}
\end{equation}
where $d_{se} = d_3 + d_4$ is the distance from the shoulder to the elbow and $d_{ew} = d_5 + d_6$ is the distance from the elbow to the wrist. In (\ref{eq: q_4}), the binary redundancy parameter $\mathbf{j}_e \in \{-1,1\}$ distinguishes between the elbow-up and the elbow-down configuration. 

The constellation of the shoulder, elbow, and wrist position forms two triangles of which the sides have a constant length for a given end-effector pose. Further, these three points and the two triangles lie on a plane, denoted as arm plane, which can be rotated around the vector $\mathbf{p}_{sw}$ resulting in two cones, see Fig. \ref{fig: IK visual}. 
Thereby, the elbow position ${p}_e$ always stays on the perimeter of the cone bases. 
As a result, the robot can perform self-motions by moving the elbow on this perimeter. 
To this end, an arm angle $\varphi$ is introduced as a redundancy parameter, referring to the angle between a reference arm with the special configuration $q_{3,n} = 0$ (red lines in Fig. \ref{fig: IK visual}), and the actual arm plane (blue lines in Fig. \ref{fig: IK visual}). 
Here and in the following, the index $n$ refers to the reference arm configuration. 
The actual elbow orientation $\mathbf{R}_0^4$ is equivalent to rotating the orientation of the reference elbow orientation $\mathbf{R}_{0,n}^{4}$ about the shoulder-wrist vector $\mathbf{p}_{sw}$ by $\varphi$, i.e. 
\begin{equation}
\mathbf{R}_0^4 = \mathbf{R}_\varphi \mathbf{R}_{0,n}^{4},
\label{eq: R_0_4}
\end{equation} 
with Rodrigues' formula \cite{lynch2017modern}
\begin{equation}
\mathbf{R}_{\varphi} = \mathbf{I}_{3\times3} + [\mathbf{p}_{sw}]_{\times}\sin \varphi   + [\mathbf{p}_{sw}]_{\times}^2(1-\cos \varphi)\:\:,
\end{equation}
where $\mathbf{I}_{3\times3}$ is the identity matrix, and $[\mathbf{a}]_\times$ denotes the skew-symmetric matrix of the vector $\mathbf{a}$.
Since $q_4$ remains unchanged between the reference arm configuration and the actual arm configuration for a given end-effector pose, (\ref{eq: R_0_4}) leads to 
\begin{subequations} \label{eq: R_virtual}
\begin{align}
\label{R_3_4}
\mathbf{R}_3^4 &= \mathbf{R}_{3,n}^4 \\
\label{eq: R_0_3}
\mathbf{R}_0^3 &= \mathbf{R}_\varphi \mathbf{R}_{0,n}^{3} \\
\label{eq: R_4_7}
\begin{split}
\mathbf{R}_4^7 &= (\mathbf{R}_0^3 \mathbf{R}_3^4)^\mathrm{T}\mathbf{R}_{0,d}^7  \\
&= (\mathbf{R}_\varphi \mathbf{R}_{0,n}^3 \mathbf{R}_{3,n}^4)^\mathrm{T}\mathbf{R}_{0,d}^7
\end{split}
\end{align}
\end{subequations} 

\begin{figure}
    \centering
    \def\svgwidth{0.79\columnwidth}
    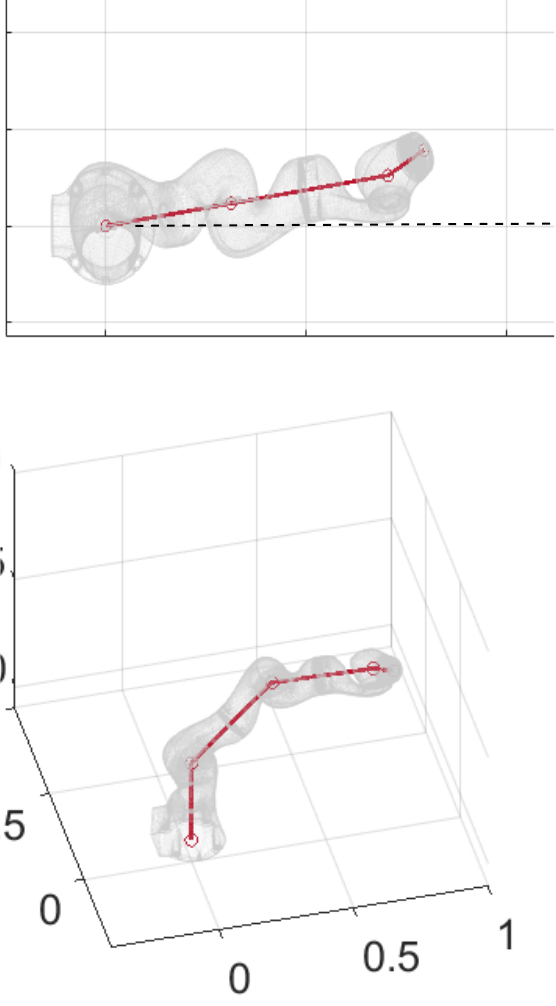
    \caption{The redundant manipulator ($q_{3,n} = 0$) in the $xy$-plane (a) and in the 3D $xyz$-plane (b). The shoulder, elbow, and wrist positions are colinear in (a).}%
\label{figure: IK visual helper}%
\end{figure}

Note that $\mathbf{R}_{0,n}^{3}$ depends only on the joint angles $q_{1,n}$ and $q_{2,n}$, since $q_{3,n} = 0$ in the reference configuration. The joint angles $q_{1,n}$ and $q_{2,n}$, shown in Fig. \ref{figure: IK visual helper}, are simply found as
\begin{subequations}
\begin{align}
q_{1,n} &= \arctantwo({p}_{sw,x},{p}_{sw,y}) \\
q_{2,n} &= \arctantwo\Bigl(\sqrt{({p}_{sw,x})^2+({p}_{sw,y})^2},{p}_{sw,z}\Bigl) + \gamma \:\:,
\end{align}
\label{eq: q_1 q_2}
\end{subequations}
with $\mathbf{p}_{sw}^\mathrm{T} = [{p}_{sw,x},{p}_{sw,y},{p}_{sw,z}]$, and 
\begin{equation*}
\gamma = j_e\arccos \left(\dfrac{d_{se}^2 + \lvert\mathbf{p}_{sw}\rvert^2 - d_{ew}^2}{2 d_{se}\lvert\mathbf{p}_{sw}\rvert} \right).
\end{equation*}
Note that $\mathbf{R}_0^3$ and $\mathbf{R}_4^7$ can be directly computed using (\ref{eq: R_0_3}), (\ref{eq: R_4_7}) and Tab. \ref{table: coordinate transformation}. 
Analytically, the rotation matrices $\mathbf{R}_0^3$ and $\mathbf{R}_4^7$ result from Tab. \ref{table: coordinate transformation} in the form
\begin{subequations}
\label{eq: anal-R}
\begin{align}
\mathbf{R}_0^3  &= \begin{bmatrix}
* & *& \cos q_1 \sin q_2 \\
* & *& \sin q_1 \sin q_2 \\
-\sin q_2 \cos q_3 & \sin q_2 \sin q_3 & \cos q_2
\end{bmatrix} \\
\mathbf{R}_4^7  &= \begin{bmatrix}
* & *& \cos q_5 \sin q_6 \\
-\sin q_6 \cos q_7 & \sin q_6 \sin q_7 & \cos q_6\\
* & *& -\sin q_5 \sin q_6 
\end{bmatrix},
\end{align}
\end{subequations}
where the elements written as $*$ are omitted for brevity. From (\ref{eq: anal-R}), the joint angles of the redundant manipulator are computed in a straightforward manner
\begin{equation}
\begin{aligned}
q_1 &= \arctantwo(\mathbf{R}_{0}^3[2,3],\mathbf{R}_0^3[1,3]) \\
q_2 &= j_s\arccos(\mathbf{R}_0^3[3,3]) \\
q_3 &= \arctantwo(\mathbf{R}_0^3[3,2],-\mathbf{R}_0^3[3,1]) \\
q_5 &= \arctantwo(-\mathbf{R}_4^7[3,3],\mathbf{R}_4^7[1,3]) \\
q_6 &= j_w\arccos(\mathbf{R}_4^7[2,3]) \\
q_7 &= \arctan2(\mathbf{R}_4^7[2,2],-\mathbf{R}_4^7[2,1])\:, \\
\end{aligned}
\label{eq: IK sol q_res}
\end{equation}
where $j_s,j_w \in \{-1,1\}$ are the remaining binary redundancy parameters and $\mathbf{R}[i,j]$ is the matrix element of the $i$-th row and $j$-th column of $\mathbf{R}$.

In summary, the parameterization of the inverse kinematics solution uses the three binary variables $\mathbf{j}_{c}^\mathrm{T} = [j_{s},j_{e},j_w]$ and the arm angle $\varphi$ in (\ref{eq: q_4}) and (\ref{eq: IK sol q_res}) as redundancy parameters to determine a unique joint configuration $\mathbf{q}$ for a desired end-effector pose $\mathbf{T}_{0,d}^{e}$. In Fig. \ref{fig: IK visual}, the blue lines illustrate a possible robot configuration with $\mathbf{j}_c = [1,-1,1]^\mathrm{T}$ that is rotated by $\varphi =$ \SI{95}{\degree} from the reference arm plane, drawn with red lines. To this end, by combining (\ref{eq: q_4}) and (\ref{eq: IK sol q_res}), the unique analytical inverse kinematics of the KUKA LBR iiwa 14 reduces to the compact form
\begin{equation}\label{eq: AIK}
\mathbf{q} = \mathrm{AIK}(\mathbf{T}_{0,d}^e,\mathbf{j}_c,\varphi),
\end{equation}
with the redundancy parameters $\mathbf{j}_c \in \{1,-1\}^3$ and $\varphi \in [0,2\pi]$.  
\subsection{Point-to-point trajectory optimization}
\label{subsection: Point-to-point trajectory optimization}
In the point-to-point (PTP) trajectory planning, a desired trajectory $\bm{\xi}^{*}(t) = [\mathbf{x}^*(t),\mathbf{u}^*(t)]^{\mathrm{T}}, \:t\in [t_0,t_F]$ for the robotic system (\ref{eq: state space 2}) is planned from an initial configuration $\bm{\xi}^*(t_0)= [\mathbf{x}_{t_0},\mathbf{u}_{t_0}]^\mathrm{T}$ to a 
target configuration $\bm{\xi}^*(t_F) = [\mathbf{x}_{t_F},\mathbf{u}_{t_F}]^\mathrm{T}$. The target configuration has to satisfy the forward kinematics relation for the desired end-effector pose $\mathbf{T}_{0,d}^{e}$, see (\ref{eq: FK})
\begin{equation}
\mathbf{T}_{0,d}^{e} - \mathrm{FK}(\mathbf{q}_{t_F}) = \mathbf{0}
\label{eq: FK condition}
\end{equation}
Without loss of generality, the initial time $t_0$ is chosen as $t_0=0$. Furthermore, the target configuration is assumed to be a stationary point $\mathbf{x}_{t_F}^\mathrm{T} = [\mathbf{q}_{t_F}^\mathrm{T},\mathbf{0}^\mathrm{T}]$. 

The PTP trajectory planning is formulated as optimization problem using the direct collocation method, see, e.g., \cite{betts2010practical}, by discretizing the trajectory $\bm{\xi}(t), t\in [0,t_F]$, with $N+1$ grid points and solving the resulting static optimization problem
\begin{subequations}\label{Eq: discrete}
\begin{align}
\label{Eq: discrete a}
 \min_{\bm{\xi}^*} &\: J(\bm{\xi}) = t_{F} + \dfrac{1}{2}h\sum_{k=0}^{N} \mathbf{u}_{k}^\mathrm{T} \mathbf{R} \mathbf{u}_{k} \\
\label{Eq: discrete b}
\text{s.t.} \:\: &\mathbf{x}_{k+1} - \mathbf{x}_{k} = \dfrac{1}{2}h
\begin{bmatrix}
\dot{\mathbf{q}}_{k+1} + \dot{\mathbf{q}}_{k} \\
\mathbf{u}_{k+1} + \mathbf{u}_k
\end{bmatrix} \\
\label{Eq: discrete c}
& \mathbf{x}_0 = \mathbf{x}_{t_0}, \:\: \mathbf{x}_N= \mathbf{x}_{t_F} \\
\label{Eq: discrete d}
& \underline{\mathbf{x}} \leq \mathbf{x}_{k}  \leq \overline{\mathbf{x}} \\
\label{eq: costly}
&\underline{\bm{\tau}} \leq {\mathbf{M}}(\mathbf{q}_{k})\mathbf{u}_{k} + {\mathbf{C}}(\mathbf{q}_{k},\dot{\mathbf{q}}_{k})\dot{\mathbf{q}}_{k} + {\mathbf{g}}(\mathbf{q}_{k}) \leq \overline{\bm{\tau}}\:\\
&k=0,\dots,N\:\nonumber
\end{align}
\end{subequations}
for the optimal trajectory 
\begin{equation}
(\bm{\xi}^*)^\mathrm{T} = [t_{F}^*,(\mathbf{x}_{0}^*)^\mathrm{T}, \dots, (\mathbf{x}_{N}^*)^\mathrm{T},(\mathbf{u}_{0}^*)^\mathrm{T}, \dots, (\mathbf{u}_{N}^*)^\mathrm{T}],
\label{eq: optimal val}
\end{equation} 
with the time step $h= t_F/N$. Note that the final time $t^*_{F}$ in (\ref{eq: optimal val}) denotes the optimal duration of the trajectory from the initial state $\mathbf{x}_{t_0}$ to the target state $\mathbf{x}_{t_F}$. 
In addition, $\mathbf{R}$ is a positive definite weighting matrix for the input $\mathbf{u}$ which also weighs the tradeoff between the cost of the duration and the smoothness of the trajectory. The system dynamics (\ref{eq: state space 2}) is approximated by the trapezoidal rule in (\ref{Eq: discrete b}). Moreover, $\underline{\mathbf{x}}$ and $\overline{\mathbf{x}}$ in (\ref{Eq: discrete d}) denote the symmetric lower and upper bounds of the state, respectively, and (\ref{eq: costly}) considers the upper and lower torque limit $\overline{\bm{\tau}}$ and $\underline{\bm{\tau}}$.

It should be noted that (\ref{eq: costly}) is a computationally expensive inequality constraint, mainly because of the large expressions in the Coriolis matrix ${\mathbf{C}}(\mathbf{q},\dot{\mathbf{q}})$. Indeed, the Coriolis matrix is often neglected in industrial applications \cite{binder1986distributed,vu2021fast}. To still consider the influence of the Coriolis matrix ${\mathbf{C}}(\mathbf{q},\dot{\mathbf{q}})$ for the torque limits, the range of values of the term ${\mathbf{C}}(\mathbf{q},\dot{\mathbf{q}})\dot{\mathbf{q}}$ is investigated for the KUKA LBR iiwa 14 R820 using a Monte Carlo simulation. In this simulation, $10^8$ uniformly distributed random state vectors $\mathbf{x}$ are selected from the admissible operating range, see Tab. \ref{table: constraints}. This simulation shows that the values of ${\mathbf{C}}\dot{\mathbf{q}}$ are between $\overline{\mathbf{c}}^\mathrm{T} = [6,8,3,4,1,1,0. 1]$ \SI{}{\newton\meter} and $\underline{\mathbf{c}}^{\mathrm{T}} = -[6,7,3,4,1,1,0.1]$ \SI{}{\newton\meter}, which is much smaller than the torque limits of the motor. Although the influence of the Coriolis matrix on the dynamics of the overall system is not significant, it is still advantageous to consider these physical limits in the optimization problem (\ref{Eq: discrete}). To this end, the costly inequality condition (\ref{eq: costly}) is replaced by
\begin{equation}
\underline{\bm{\tau}} - \underline{\mathbf{c}} \leq ({\mathbf{M}}(\mathbf{q}_{k})\mathbf{u}_{k} + {\mathbf{g}}(\mathbf{q}_{k})) \leq \overline{\bm{\tau}} - \overline{\mathbf{c}}\: .
\label{eq: no-costly}
\end{equation}
The optimal trajectory is computed by solving the static optimization problem (\ref{Eq: discrete a})-(\ref{Eq: discrete d}) and (\ref{eq: no-costly}) using Interior Point OPTimize (IPOPT), an open-source package based on the interior point method (IPM) for large-scale nonlinear programming, see, e.g., \cite{wachter2006implementation}. 
\subsection{Optimal target configuration $\mathbf{q}_{t_F}$}
\label{subsection: optimal target conf}
In this section, the optimal choice for the target configuration $\mathbf{q}_{t_F}$ is discussed. The inverse kinematics for a redundant robot does not yield a unique joint configuration $\mathbf{q}_{t_F}$, as presented in Section \ref{subsection: Analytic inverse kinematics}. Moreover, choosing an unsuitable target configuration $\mathbf{q}_{t_F}$ may cause the trajectory optimization (\ref{Eq: discrete}) to fail or deliver poor results. 

For redundant robots, there is an infinite number of joint configuration solutions $\mathbf{q}_{t_F}$ for a desired end-effector pose $\mathbf{T}_{0,d}^{e}$. Therefore, two criteria for selecting the best inverse kinematics solution, i.e. the manipulability and closeness, are introduced in the following and an optimization problem is formulated.  

First, the manipulability $m(\mathbf{q})$ \cite{yoshikawa1985manipulability} is the most popular index used to measure the dexterity of a robot for a specific joint configuration $\mathbf{q}$. It is defined as
\begin{equation}
m(\mathbf{q}) = \sqrt{\det \left( \mathbf{J}(\mathbf{q})\mathbf{J}^{\mathrm{T}}(\mathbf{q}) \right)}\:\:,
\label{eq: manipulability-classical}
\end{equation}
where the geometric manipulator Jacobian $\mathbf{J}(\mathbf{q})$ takes the form
\begin{equation}
\mathbf{J}(\mathbf{q}) = \begin{bmatrix}
\mathbf{J}_v(\mathbf{q})\\
\mathbf{J}_\omega(\mathbf{q})
\end{bmatrix}
= \begin{bmatrix}
\dfrac{\partial \mathbf{p}_t(\mathbf{q})}{\partial \mathbf{q}}\\
\dfrac{\partial \bm{\omega}_t(\mathbf{q})}{\partial \mathbf{q}}
\end{bmatrix}.
\label{eq: geometric jacobian}
\end{equation}
In (\ref{eq: geometric jacobian}), $\bm{\omega}_t$ is the angular velocity of the end effector described in the frame $\mathcal{O}_0$, which is computed by
\begin{equation}
[\bm{\omega}_t]_{\times} = \dot{\mathbf{R}}_0^7(\mathbf{q})\Big(\mathbf{R}_0^7(\mathbf{q}\Big)^\mathrm{T}.
\end{equation}
To reduce the computational burden of (\ref{eq: manipulability-classical}) due to the computation of the determinant, an analytical expression of the manipulability is derived, which is given in the appendix. 

Second, to consider the closeness between the inverse kinematics solution $\mathbf{q}$ and the initial joint configuration $\mathbf{q}_0$ of the robot, the $L_\infty$-norm $\lVert . \rVert_{\infty}$ is employed to find the largest deviation between these two joint space configurations. Here, the closeness is given by
\begin{equation}
c(\mathbf{q}) = \lVert\mathbf{q}_{0} - \mathbf{q} \rVert_{\infty},
\label{eq: closeness}
\end{equation}
where $\mathbf{q}_{0}$ is the initial joint configuration of the initial state $\mathbf{x}_{0} = [\mathbf{q}_{0}^\mathrm{T},\mathbf{0}]$. 

\begin{figure}
    \centering
    \def\svgwidth{1\columnwidth}
    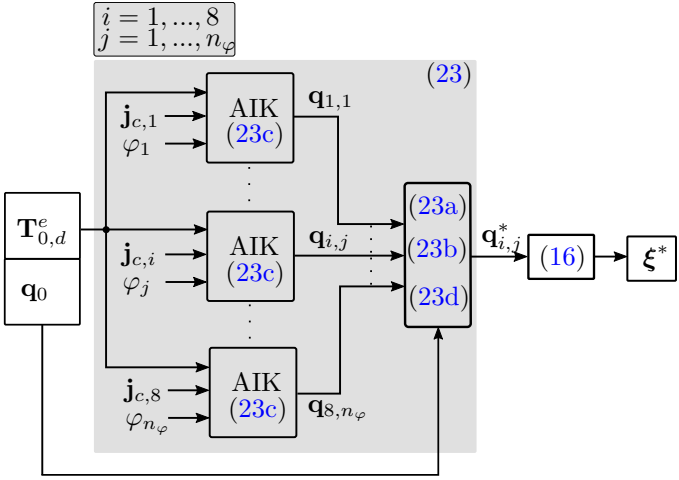
    \caption{\textcolor{black}{Block diagram of the optimization problem (\ref{eq: manipulability and closeness}) and (\ref{Eq: discrete})}. 
    }%
    \label{fig: outline_method}%
\end{figure}

Next, the two criteria (\ref{eq: manipulability-classical}) and (\ref{eq: closeness}) are considered in an optimization problem to choose the best target configuration $\mathbf{q}_{t_F}$ for a given target pose $\mathbf{T}_{0,d}^e$. To solve this problem, according to (\ref{eq: AIK}), the redundancy parameters of the inverse kinematics $\mathbf{j}_c$ and $\varphi$ have to be determined.
Since there are three binary redundancy parameters in $\mathbf{j}_c$, $2^3=8$ different values are contained in the set $\mathcal{X}_{\mathbf{j}_c} = \{\mathbf{j}_{c,i}\: \lvert\: i = 1,...,8\}$. Additionally, the arm angle $\varphi \in [0,2\pi]$ is equidistantly discretized with the grid points
\begin{equation*}
\mathcal{X}_{\varphi} = \bigg\{j\dfrac{2\pi}{n_\varphi} \:\Big\lvert \:\:j=1,...,n_\varphi\bigg\}\:\:.
\end{equation*}
The following optimization problem is solved to find the best target configuration $\mathbf{q}_{t_F} = \mathbf{q}^*_{i,j}$ as well as its corresponding redundancy parameters $\mathbf{j}_c^*$ and the virtual angle $\varphi^*$ for the desired pose $\mathbf{T}_{0,d}^{e}$
\begin{subequations}\label{eq: manipulability and closeness}
\begin{align}
\label{eq: mani main} & \argmin_{\mathbf{q}_{i,j}^*,\mathbf{j}_c^*,\mathbf{\varphi}^*} \bigg\{J_{IK}(\mathbf{q}_0,\mathbf{q}_{i,j})\bigg\}_{
\begin{matrix}
i\in \{1,...,8\} \\
j \in \{1,...,n_\varphi\}
\end{matrix}
}\\
\label{eq: mani cost}
&\text{s.t.} \:\: J_{IK}(\mathbf{q}_0,\mathbf{q}_{i,j}) = \dfrac{\omega_{m}}{m(\mathbf{q}_{i,j})}  + \omega_{c}c(\mathbf{q}_{0},\mathbf{q}_{i,j}) \\
\label{eq: mani AIK}
&\:\: \:\: \:\: \:\: \mathbf{q}_{i,j} = \mathrm{AIK}(\mathbf{T}_{0,d}^{e},\mathbf{j}_{c,i},\varphi_j) \\
\label{eq: mani constraint}
&\:\: \:\: \:\: \:\: \underline{\mathbf{q}} \leq \mathbf{q}_{i,j}  \leq \overline{\mathbf{q}}\:\:,
\end{align}
\end{subequations}
with the user-defined weighting parameters $\omega_m > 0$ and $\omega_c >0$. 
\textcolor{black}{
To compute an optimal trajectory $\bm{\xi}^*$ from the current robot configuration, represented by $\mathbf{q}_0$, to the given desired pose $\mathbf{T}_{0,d}^e$, the optimization problem (\ref{eq: manipulability and closeness}) is solved first to obtain the optimal solution $\mathbf{q}_{i,j}^*=\mathbf{q}_{t_F}$. Then, this optimal robot target configuration $\mathbf{q}_{t_F}$ is used in the PTP trajectory optimization (\ref{Eq: discrete}). 
The block diagram of this process is illustrated in Fig.~\ref{fig: outline_method}.
}
\section{Framework for learning redundancy parameters}
\label{section: Framework for learning analytic inverse kinematics parameters}
\begin{figure*}
    \centering
    \def\svgwidth{2\columnwidth}
    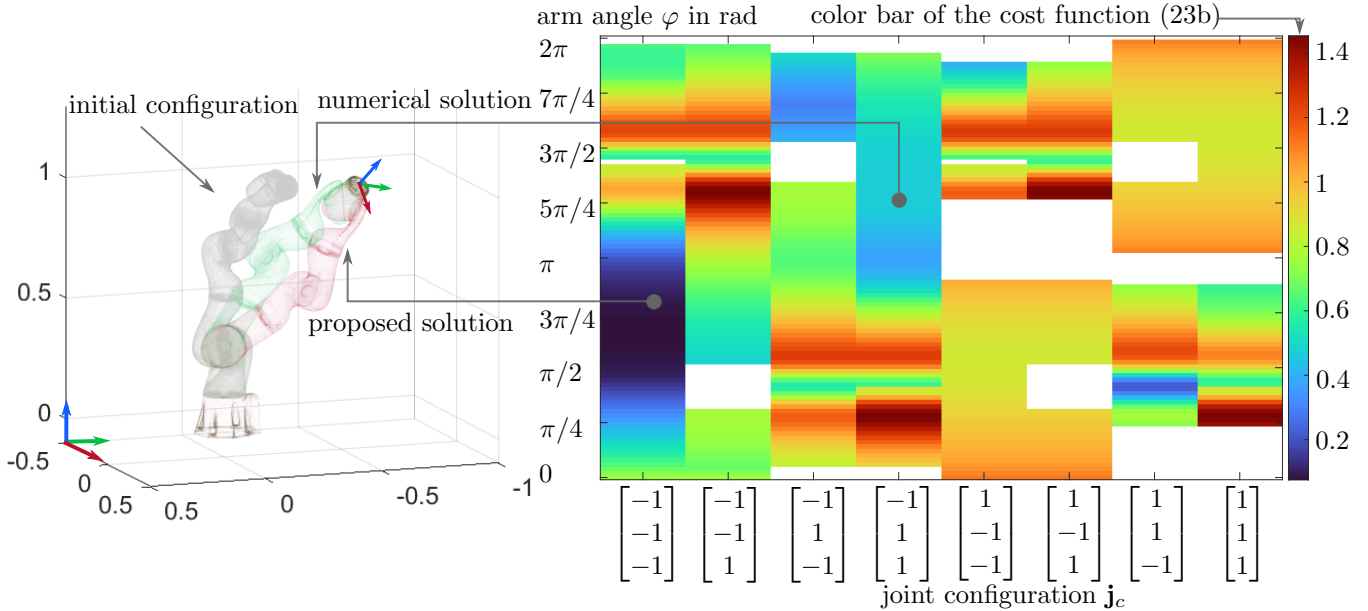
    \caption{The color map of the cost function (\ref{eq: mani cost}) for the initial configuration of the robot $\mathbf{q}_0$ (in gray color) and the position of the end effector (the RGB triad) is shown on the right-hand side of this figure. The white color regions depict infeasible joint configurations. The robot target configuration $\mathbf{q}_{t_F}$ computed by the proposed NN is shown in red color on the left-hand side. This achieves a very small value of the cost function (\ref{eq: mani cost}) ($\approx 0.0717$). The target robot configuration calculated using the numerical method \cite{buss2004introduction} is depicted in green. The cost of this configuration is approximately $0.54$.}%
    \label{fig: RAL_sim_1}%
\end{figure*}
The cost function (\ref{eq: mani cost}) and the inverse kinematics (\ref{eq: mani AIK}) are nonlinear and discontinuous functions with many local minima, which is illustrated on the right-hand side of Fig. \ref{fig: RAL_sim_1} for an example joint configuration $\mathbf{q}$. Therefore, to find the global optimum, the optimization problem (\ref{eq: manipulability and closeness}) has to be solved by exhaustive search, which is a time-consuming process since (\ref{eq: mani AIK}) has to be evaluated $8n_\varphi$ times, see Fig. \ref{fig: outline_method}. To significantly reduce the computational effort for this step, a neural network (NN) is presented in this section to quickly determine the joint configuration $\mathbf{j}_c^*$ and narrow down the search space for the arm angle $\varphi^*$ for a desired end-effector pose $\mathbf{T}_{0,d}^e$ and the given initial configuration $\mathbf{q}_0$.

First, the generation of the database to train the NN for learning the redundancy parameters $\mathbf{j}_c$ and $\varphi$ is introduced. Then, the network architecture of this NN is presented in the next step.  

\subsection{Database generation}
\begin{figure}
    \centering
    \def\svgwidth{0.7\columnwidth}
    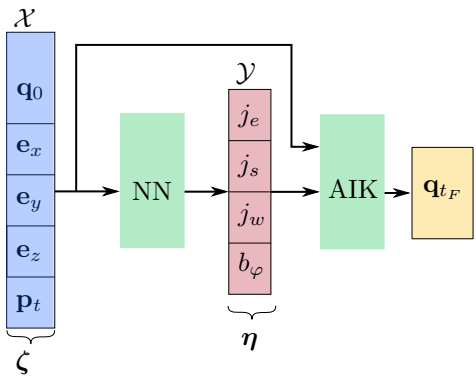
    \caption{Overview of the proposed framework for learning the redundancy parameters.}%
    \label{fig: IK overview learning}%
\end{figure}

\begin{figure*}
    \centering
    \def\svgwidth{2\columnwidth}
    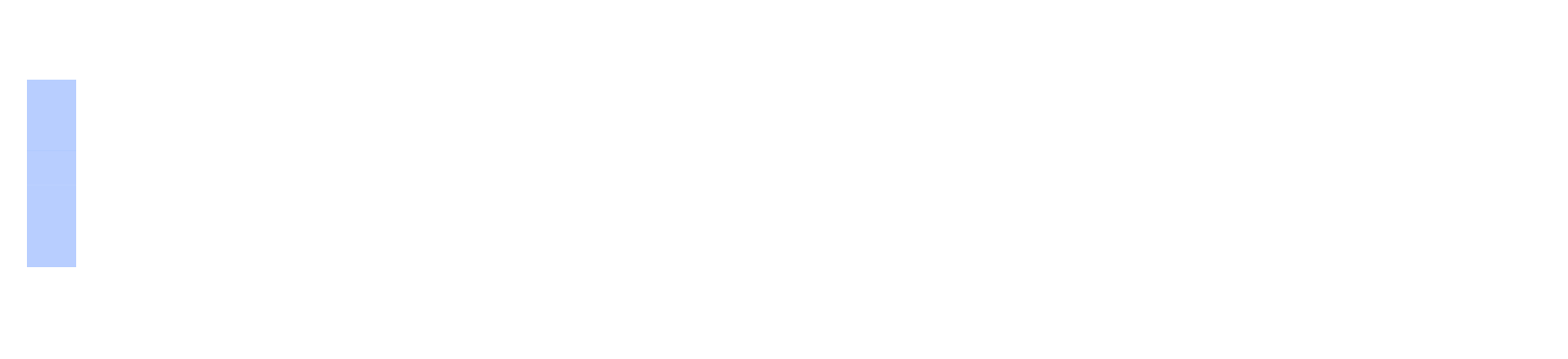
    \caption{Architecture of the proposed NN for learning the joint configuration $\mathbf{j}_c$ and the bin index $b_\varphi$ of the virtual angle $\varphi$.}%
    \label{fig: IK learning scheme}%
\end{figure*}
\label{subsection: Database generation}

For the database generation, $N_p$ pairs of robot initial joint configurations $\mathbf{q}_{0,k}$ and corresponding feasible desired poses $\mathbf{T}_{0,d}^{e,k}$ are randomly selected from a uniform random distribution in the admissible ranges and are stored in the set $\mathcal{X}=\{\bm{\zeta}_k\}=\{\mathbf{q}_{0,k},\mathbf{T}_{0,d}^{e,k}\:\big\lvert\:{k=1,...,N_p}\}$. 
 For each pair of $\mathbf{q}_{0,k}$ and $\mathbf{T}_{0,d}^{e,k}$, the optimization problem (\ref{eq: manipulability and closeness}) is solved by an exhaustive search to find the global optimum redundancy parameters $\mathbf{j}_{c}^*$ and $\varphi^*$ as well as the target configuration $\mathbf{q}_{t_F}^*$. The redundancy parameters are stored in the set  $\mathcal{Y}=\{\bm{\eta}_k\}=\{\mathbf{j}_{c,k},\varphi_k\:\lvert\:k=1,...,N_p\}$. The database $\mathcal{D} = (\mathcal{X},\mathcal{Y})$ comprises both sets $\mathcal{X}$ and $\mathcal{Y}$. 
Elements from the set $\mathcal{X}$ are the input to the NN and elements from the set $\mathcal{Y}$ are the corresponding labeled outputs, see Fig. \ref{fig: IK overview learning}. 

The input data in the set $\mathcal{X}$ contain redundant data due to the constant bottom row in the desired pose $\mathbf{T}_{0,d}^{e,k}$, see (\ref{eq: T_0_e}). Therefore, 
only the three basis vectors $\mathbf{e}_{x,k}$, $\mathbf{e}_{y,k}$, and $\mathbf{e}_{z,k}$, of $\mathbf{R}_0^{e,d} = [\mathbf{e}_{x,k},\mathbf{e}_{y,k}, \mathbf{e}_{z,k}]$ and the position of the end effector $\mathbf{p}_{t,k}$ are considered in the set $\mathcal{X}$. Thus, the input set $\mathcal{X}$ is re-arranged in the form
\begin{equation*}
\mathcal{X} = \{\bm{\zeta}_k\:\lvert\:k=1,...,N_p\},
\end{equation*}
with 
\begin{equation*}
\bm{\zeta}_k^\mathrm{T} = [(\mathbf{q}_{0,k})^\mathrm{T},(\mathbf{e}_{x,k})^\mathrm{T},(\mathbf{e}_{y,k})^\mathrm{T}, (\mathbf{e}_{z,k})^\mathrm{T},(\mathbf{p}_{t,k})^\mathrm{T}] \in \mathbb{R}^{19}
\end{equation*}


Since (\ref{eq: mani cost})-(\ref{eq: mani AIK}) are discontinuous nonlinear functions, see Fig.~\ref{fig: RAL_sim_1}, a complex NN is required to approximate these functions. However, the training and prediction time of such a neural network is very long, making it impossible to be implemented in real time. 
\textcolor{black}{Thus, instead of directly predicting the virtual angle $\varphi$, only the range of this angle, denoted by the bin index $b_{\varphi} \in \{1,...,n_b\}$ with the total number of bins $n_b$, is predicted. 
This way, the value of the bin index $b_\varphi$ indicates that the virtual angle $\varphi$ is in the range $\bigg[(b_\varphi-1)\dfrac{2\pi}{n_b},b_\varphi \dfrac{2\pi}{n_b}\bigg], b_\varphi=1,...,n_b$. 
This helps to reduce the complexity and to realize the proposed NN for a real-time application. 
}
Consequently, $\varphi_k$ is replaced by its bin index $b_{\varphi,k} \in \{1,...,n_b\}$ in the set $\mathcal{Y}$ resulting in
$
\mathcal{Y} = {\bm{\eta}_k}=\{\mathbf{j}_{c,k}, b_{\varphi,k}\}. 
$


\subsection{Network Architecture}
\label{subsection: Network Architecture}
The architecture of the proposed NN is shown in Fig. \ref{fig: IK learning scheme}. This NN is designed for the two sub-problems, i.e., to learn the joint configuration $\mathbf{j}_c$ and the bin index $b_\varphi$ of the arm angle $\varphi$. Note that the input of the proposed NN is $\bm{\zeta} \in \mathcal{X}$  and the output is the predicted value $\bm{\eta}^\mathrm{T} = [\mathbf{j}_c^\mathrm{T}, b_\varphi]\in \mathcal{Y}$.

First, two fully connected layers of size $32$ with a ReLU activation function \cite{li2017convergence} are utilized, as shown in Fig. \ref{fig: IK learning scheme}. Since there are 8 possibilities for choosing $\mathbf{j}_c$, a fully connected layer of size $8$ with a softmax activation function \cite{bishop2006pattern} is employed to output $\mathbf{j}_c$. The cross-entropy function is used to compute the loss between the prediction $\hat{\mathbf{j}}_{c,k}$ of the NN and the target value ${\mathbf{j}}_{c,k}$ in the form
\begin{equation}
\mathrm{L}_{\mathbf{j}_c} = \sum_{k=1}^M -\mathbf{j}^{\mathrm{T}}_{c,k}\log(\hat{\mathbf{j}}_{c,k}) + (\mathbf{1}-\mathbf{j}_{c,k})^{\mathrm{T}}\log(\mathbf{1}-\hat{\mathbf{j}}_{c,k})\:\:,
\end{equation}
where $M$ is the size of the training dataset. 

Second, the predicted $\hat{\mathbf{j}}_c$ is concatenated with the input $\bm{\zeta}$ again as a new input for the second subproblem. Similar to the first subproblem, two fully connected layers of size $32$ with a ReLU activation function are used. Subsequently, the fully connected layer of size $8$ and the softmax activation function are implemented to predict the bin index $b_\varphi$ of the arm angle $\varphi$. Again, the cross entropy function is used to compute the loss between the predicted value of the bin index $\hat{b}_\varphi$ and the target value $b_\varphi$
\begin{equation}
\mathrm{L}_{b_\varphi} = \sum_{k=1}^M -b_{\varphi,{k}}\log(\hat{b}_{\varphi,{k}}) + ({1}-b_{\varphi,{k}})\log(1-\hat{b}_{\varphi,k}).
\end{equation}

\textcolor{black}{The proposed NN is trained by using the Adam \cite{kingma2014adam} optimizer with the learning rate of $\alpha=10^{-3}$. Furthermore, the $\mathrm{L}_2$ regularization \cite{kingma2014adam} with $\lambda = 10^{-6}$ is added to both loss functions $\mathrm{L}_{\mathbf{j}_c}$ and $\mathrm{L}_{b_\varphi}$. This helps to avoid overfitting \cite{murphy2012machine}.}


\section{Results}
\label{section: simulation results}
The simulation results presented in this section are obtained on a computer with a 3.4 GHz Intel Core i7-10700K and 32 GB RAM. The generated database with $N_p = 10^8$ pairs described in Section \ref{subsection: Database generation} is randomly shuffled and divided into $3$ subsets, i.e., training dataset, validation dataset, and test dataset, which are partitioned as $80\%$, $10\%$, and $10\%$ of the generated database, respectively. To speed up the computing time of database generation, C++ code is generated for (\ref{eq: manipulability and closeness}) using \textsc{MATLAB} coder in \textsc{MATLAB} R2021b. Additionally, the analytical expression of the manipulability in the appendix (\ref{eq: mani anal}) is utilized. The remaining parameters are chosen as $n_\varphi=100$ and $n_b=8$. For the database generation, the average computing time of (\ref{eq: manipulability and closeness}) for a given pose and initial joint configuration is approximately \SI{1.5}{\milli\second}. Since $n_\varphi$ in (\ref{eq: manipulability and closeness}) is set to $100$, the average computing time of the analytical inverse kinematics expression in (\ref{eq: mani AIK}) is approximate \SI{1.8}{\micro\second}. 

\subsection{Statistical information on training the proposed NN}
The proposed NN is trained using the open-source software package Keras \cite{gulli2017deep}. To reduce the training time, the CUDA cores of an Nvidia GeForce RTX 3070 are employed. During training, the mini-batch size is set to $2000$ and the training data is reshuffled in each epoch. 

\begin{figure}
    \centering
    \includegraphics[width=0.48\textwidth]{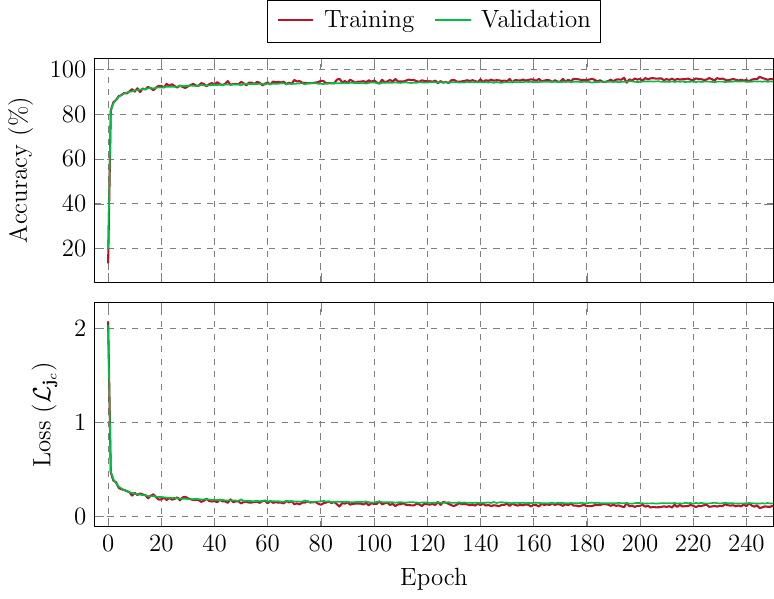}
    \caption{Training and validation accuracy of the joint configuration $\mathbf{j}_c$.}%
    \label{fig: joint configuration learning}%
\end{figure}

Fig. \ref{fig: joint configuration learning} shows that the learning accuracy for the joint configuration $\mathbf{j}_c$ of the training dataset and the validation dataset after $500$ epochs reaches $96.62\%$ and $95.76\%$, respectively. Also, the corresponding values of the loss function $\mathrm{L}_{\mathbf{j}_c}$ decreased to $0.1034$ and $0.1264$, respectively. To further validate the training result, the accuracy of the test dataset with the trained parameters of the proposed NN is approximately $96.49\%$. 

\begin{figure}
    \centering
    \includegraphics[width=0.48\textwidth]{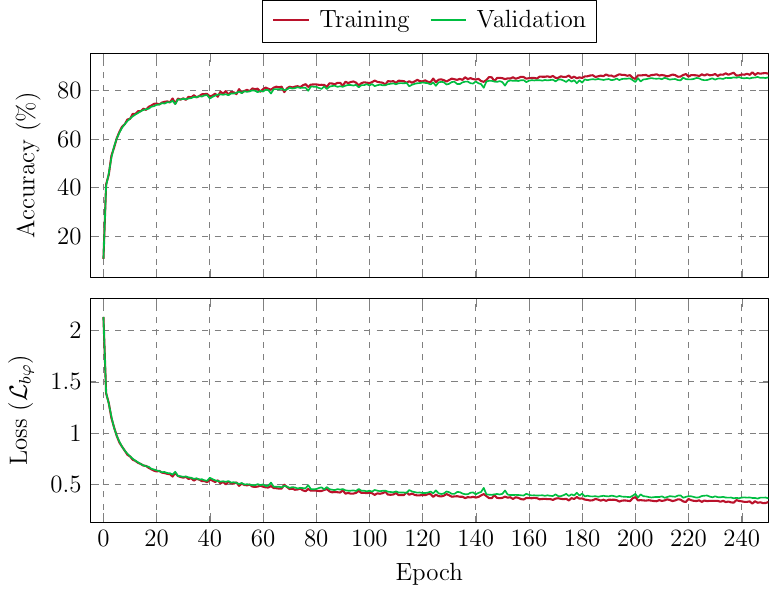}
    \caption{Training and validation accuracy of the bin index $b_\varphi$.}%
    \label{fig: angle learning}%
\end{figure}

Fig. \ref{fig: angle learning} shows the accuracy of the training dataset and the validation dataset with respect to the bin index $b_\varphi$ of the arm angle $\varphi$. Note that the resulting accuracy is approximately $85.57\%$ for the training dataset and $85.12\%$ for the validation dataset. The values of the loss function $\mathrm{L}_{b_\varphi}$ are approximately $0.32$ and $0.38$ for the two datasets. To verify the trained parameters of the proposed NN, a consistent accuracy of $84.78\%$ is reported for the test dataset. 

For further validation, the proposed NN is compared to the performance of four well-known algorithms, i.e., naive Bayes classifier \cite{hastie2009elements}, discriminant analysis classifier \cite{tharwat2016linear}, binary decision tree classifier \cite{loh2002regression} and $k$-nearest neighbor classifier \cite{jiang2007survey}. 
Similar to the proposed NN, each classifier takes the input $\bm{\zeta}\in\mathcal{X}$ and outputs the prediction $\bm{\eta}\in\mathcal{Y}$. 
\begin{table}
\caption{Performance of the prediction with different algorithms}
\label{table: performances}
\begin{center}
\scalebox{0.94}{
\begin{tabular}{
SSS[table-format=2.2]SS[table-format=2.2]}
\hline
{Classifier} &  {\makecell{Acc. $\mathbf{j}_c$\\ \%}} & {\makecell{Acc. $b_\varphi$ \\ \%}}   & {\makecell{Time \\ \SI{}{\micro\second}}} &{\makecell{Memory \\ MB}}\\ 
\hline
{\makecell{Naive Bayes \\ \cite{hastie2009elements}}} & 57.6 & 38.2 &  1.1 & 76.8\\  
\hline
{\makecell{Discriminant \\ Analysis \cite{tharwat2016linear}}} & 65.1 & 40.6 &  1.23  & 70.5 \\  
\hline
{\makecell{Binary Decision \\ Tree \cite{loh2002regression}}} & 77.8 & 65.5 & 0.35 & 89.6 \\  
\hline
{\makecell{$k$-Nearest \\Neighbor \cite{jiang2007survey}}} & 49.5&40.1 & 1810 & 76.8  \\  
\hline
{Proposed NN} & 96.5&84.8 &7.35 & 0.17  \\  
\hline
\end{tabular}
}
\end{center}
\end{table}
The statistical performance of the four algorithms and the proposed NN is shown in Tab. \ref{table: performances}. Among the above algorithms, the binary decision tree classifier achieves the highest prediction accuracy for $\mathbf{j}_c$ and $b_\varphi$, i.e., $77.8\%$ and $65.5\%$, respectively. 
Moreover, the average execution time of this classifier is approximately \SI{0.35}{\micro\second}, i.e., the fastest algorithm. However, the prediction accuracy of the proposed NN is still significantly higher compared to the binary decision tree classifier. Another aspect is the memory consumption, which is with $0.17$ MB much less compared to over $70$ MB for each of the four other algorithms. This is reasonable since in the proposed NN, the memory consumption is mainly used for storing the network parameters. 

The average NN execution time for the prediction of $\mathbf{j}_c$ and $b_\varphi$ is about \SI{7.35}{\micro\second}. Also, the average computing of the analytical inverse kinematics with the given $\mathbf{j}_c$ and $b_\varphi$ is about \SI{2}{\micro\second}. Thus, the proposed NN provides the possibility to compute a good IK solution with respect to the two criteria, i.e., manipulability (\ref{eq: manipulability-classical})  and closeness (\ref{eq: closeness}), in real time. 

\subsection{Performance of the proposed NN in the framework of trajectory optimization}
To verify the efficiency of the proposed NN, the example task of planning a PTP trajectory from the initial configuration
\begin{equation*}
\begin{aligned}
\mathbf{q}_0^{\mathrm{T}} &= [-1.5,-0.1,0.3,0.7,0.5,-0.6,1.4]\: \mathrm{rad},\\
\mathbf{j}_{c,0}^\mathrm{T} &= [-1,1,-1], \\
\varphi_0 &= 3.21 \:\mathrm{rad}\:,
\end{aligned}
\end{equation*}
to the target pose
\begin{equation}
\label{eq: T_0_e example}
\mathbf{T}_{0,d}^{e} = \begin{bmatrix}
0.863 & 0.262 & -0.433 & -0.55 \\
0.003 & 0.853 & 0.522 & 0.160 \\
0.505 & -0.451 & 0.735 & 1.049 \\
0 & 0 & 0 & 1
\end{bmatrix}\:\:, 
\end{equation} is considered. 

First, the comparison between the well-known damped least-squares inverse kinematics solution \cite{lynch2017modern,buss2004introduction} and the proposed algorithm is depicted in Fig. \ref{fig: RAL_sim_1}. 
On the right-hand side of Fig. \ref{fig: RAL_sim_1}, a color map of (\ref{eq: mani cost}) is depicted where the $x$-axis comprises the $8$ possible joint configurations $\mathbf{j}_c \in \mathcal{X}_{\mathbf{j}_c}$ and the $y$-axis contains the $n_\varphi=100$ arm angles $\varphi \in \mathcal{X}_{\varphi}$. 
Using the network architecture of Fig. \ref{fig: IK learning scheme} with $n_b=8$, the proposed NN takes about \SI{7.35}{\micro\second} to predict the joint configuration $\mathbf{j}_c = [-1,-1,-1]^\mathrm{T}$ and the bin $b_\varphi=3$, i.e. $\varphi \in [\pi/2,3\pi/4]$. 
\textcolor{black}{
To find the optimum value for the arm angle $\varphi$ inside the predicted bin, (\ref{eq: AIK}) and (\ref{eq: mani main}) are evaluated on an equidistant grid for $\varphi \in [\pi/2,3\pi/4]$ with $n_\varphi/n_b$ grid points. This way, the effort to solve the optimization problem (\ref{eq: manipulability and closeness}) reduces from $8n_\varphi=800$ to $n_\varphi/n_b\approx 13$ evaluations of (\ref{eq: AIK}) and (\ref{eq: mani cost}).}
\textcolor{black}{Since the analytical manipulability expression (\ref{eq: mani anal}) is used in (\ref{eq: mani cost}), the computing time of (\ref{eq: mani cost}) is approximately \SI{0.15}{\micro\second}, which is much smaller than (\ref{eq: AIK}).
Thus, the total execution time for computing the optimal target configuration $\mathbf{q}_{t_F}$ is approximately \SI{32}{\micro\second} including the computing time of (\ref{eq: AIK}) of \SI{2}{\micro\second}.}
On the other hand, the damped least-squares method in this example requires $17$ iterations to find the solution of the inverse kinematics with a tolerance of $10^{-8}$. 
The computing time of the numerical method is approximately \SI{3}{\milli\second}.

On the left-hand side of Fig. \ref{fig: RAL_sim_1}, the computed target configurations for the given desired target pose $\mathbf{T}_{0,d}^e$ are
\begin{equation}
\label{eq: q_A}
\begin{aligned}
\mathbf{q}_{t_F,A}^\mathrm{T} &= [-0.55,-0.96,-0.71,-0.78,-0.45,-0.8,1.55] \:\mathrm{rad},\\
\mathbf{j}_{c,A}^\mathrm{T}  &= [-1,-1,-1],\\
\varphi_A &= 2.17\: \mathrm{rad}
\end{aligned}
\end{equation}
 for the proposed algorithm (red color), and 
\begin{equation}
\begin{aligned}
\label{eq: q_N}
\mathbf{q}_{t_F,N}^{\mathrm{T}} &= [-0.7,-0.45,1.1,0.78,0.43,0.81,-0.82]\:\mathrm{rad},\\
\mathbf{j}_{c,N}^\mathrm{T}  &= [-1,1,1], \\
\varphi_N &= 3.8\: \mathrm{rad}
\end{aligned}
\end{equation}
for the damped least-squares method (green color). It is obvious that in this example the joint configuration solutions of the two methods $\mathbf{j}_{c,A}$ and $\mathbf{j}_{c,N}$ are different from the initial joint configuration $\mathbf{j}_{c,0}$.
The proposed solution has a slightly higher manipulability measure (\ref{eq: mani anal}) of $0.061$ compared to the manipulability measure of $0.045$ of the numerical solution. The closeness value (\ref{eq: closeness}) of the proposed solution is $1.49$ which is significantly smaller than the closeness value of $2.23$ of the numerical solution. 

To further demonstrate the effectiveness of the proposed IK approach, the two target configurations (\ref{eq: q_A}) and (\ref{eq: q_N}) are used in the trajectory optimization framework detailed in Section \ref{section: Trajectory Optimization}. 
The nonlinear optimization problem (\ref{Eq: discrete}) is solved using the interior point solver IPOPT \cite{wachter2006implementation} together with the linear solver MA27 \cite{duff2004ma57}. 
To increase the computational speed, the gradient and the numerical Hessian are computed using the BFGS method \cite{liu1989limited} and provided to the IPOPT solver. 
The trajectory in (\ref{Eq: discrete}) is discretized with $N=50$ collocation points, giving a total of $1051$ optimization variables. 
For this comparison, the same initial configuration $\mathbf{q}_0$ and two different target configurations $\mathbf{q}_{t_F,A}$ according to (\ref{eq: q_A}) and $\mathbf{q}_{t_F,N}$ according to (\ref{eq: q_N}) of the pose $\mathbf{T}_{0,d}^{e}$ from (\ref{eq: T_0_e example}) are used. 
While the computing time of the optimization (\ref{Eq: discrete}) for both target configurations is almost the same (\SI{55}{\milli\second}), the time for moving to the target configuration of the proposed algorithm $\mathbf{q}_{t_F,A}$ is \SI{3. 83}{\second} compared to \SI{4.03}{\second} of the numerical solution $\mathbf{q}_{t_F,N}$. 
Moreover, the cost function in (\ref{Eq: discrete a}) with $\mathbf{q}_{t_F,A}$ and $\mathbf{q}_{t_F,N}$ is $4.7$ and $5.03$, respectively. 

\begin{figure}
    \centering
    \def\svgwidth{1\columnwidth}
    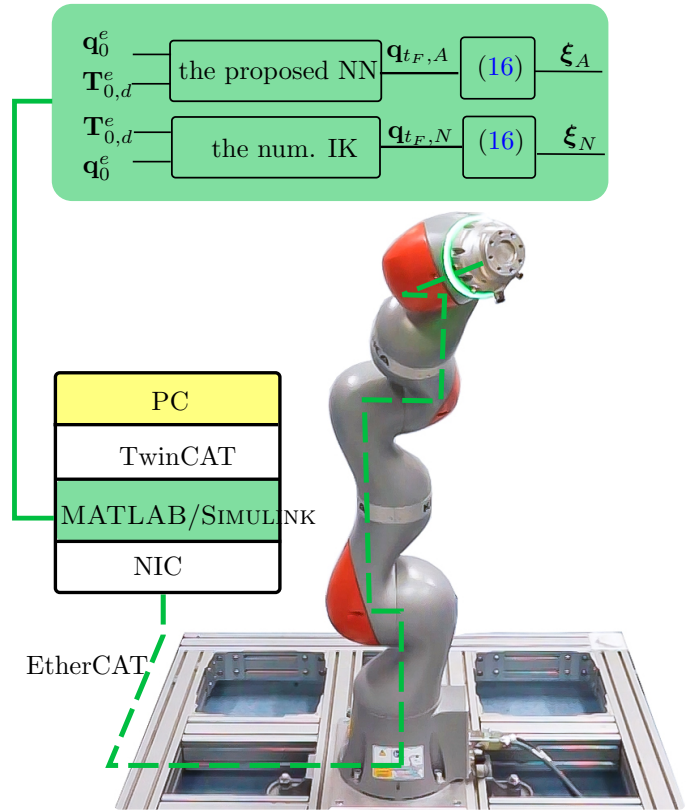
    \caption{\textcolor{black}{The experimental setup for the comparison between the proposed NN algorithm and the numerical IK method.}}%
    \label{fig: experimental setup}%
\end{figure}

\begin{figure}
    \centering
    \includegraphics[width=0.55\textwidth]{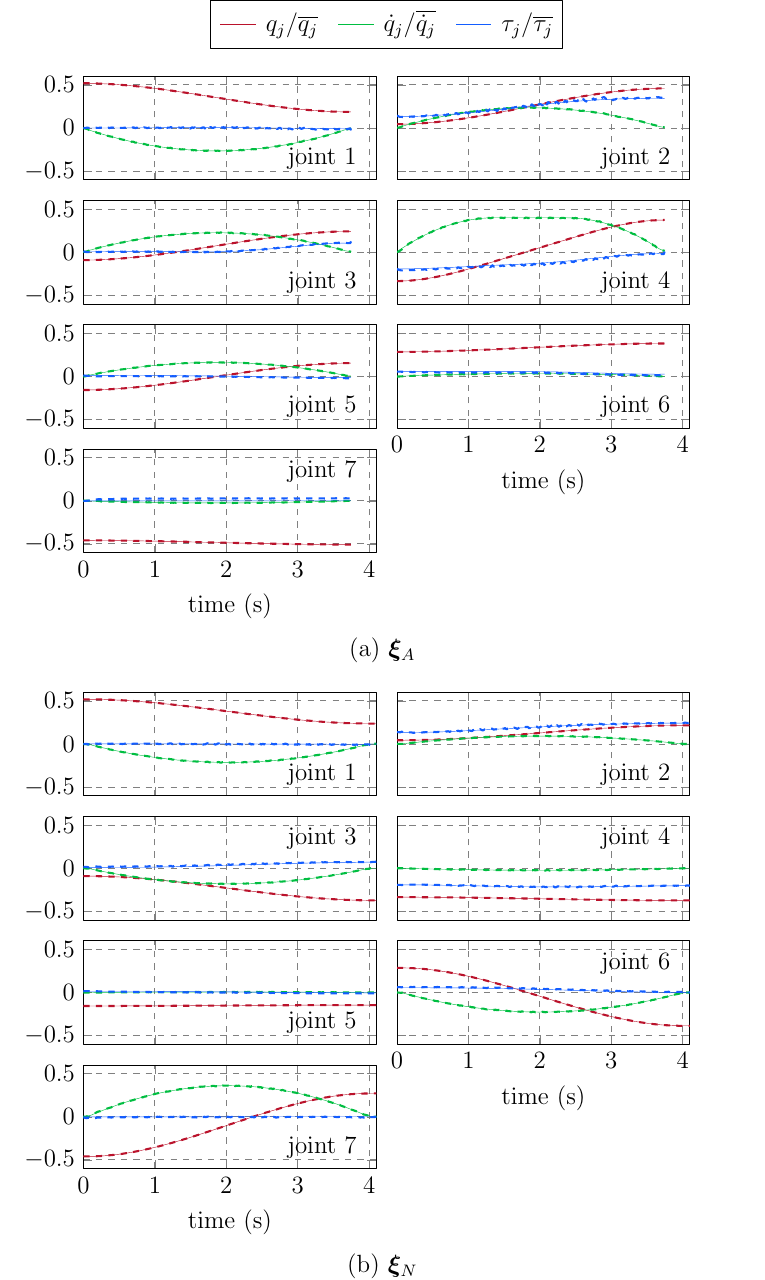}
    \caption{\textcolor{black}{Joint position, velocity, and torque for all robot axes, normalized to their respective limits, referred to with the bar symbol, for optimal trajectories $\bm{\xi}_{A}$ and $\bm{\xi}_{N}$ of the numerical IK and the proposed NN algorithm, respectively. The desired trajectories are shown as solid lines and the measured trajectories are drawn as dashed lines. 
    For safety reasons, the limits for the motor torques are $50\%$ lower than the limits in Tab.~\ref{table: constraints}.}}%
    \label{fig: time evolution exp}%
\end{figure}

\textcolor{black}{
The optimal trajectories $\bm{\xi}_{A}$ and $\bm{\xi}_{N}$ corresponding to the target configurations (\ref{eq: q_A}) and (\ref{eq: q_N}), respectively, are validated on the experimental setup depicted in Fig.~\ref{fig: experimental setup}. 
This experimental setup comprises two main components, i.e. the robot \kuka and the PC. 
The PC communicates with the robot via a network interface card (NIC) using the EtherCAT protocol. 
The computed torque controller is implemented as \textsc{MATLAB}/\textsc{Simulink} module, which is executed via the real-time automation software \textsc{Beckhoff} TwinCAT. 
The sampling time $T_s = \SI{125}{\micro\second}$ is used for the robot sensors and actuators. 
The scaled joint position, velocity, and torque for all robot axes, normalized to their respecting limits for the two optimal trajectories $\bm{\xi}_A$ and $\bm{\xi}_N$ and the corresponding measurements from the experiments are shown in Fig.~\ref{fig: time evolution exp}. 
Note that in this figure, the trajectories do not exceed the value $\pm 1$, which means that all state and input constraints in (\ref{Eq: discrete d}) and (\ref{eq: no-costly}) are respected. 
The travel time of the trajectory from the solution of the proposed NN ($\approx \SI{3.9}{\second}$) is slightly shorter than that from the numerical IK ($\approx \SI{4.1}{\second}$). 
Since the proposed NN is designed to select the configuration that is closer to the robot's initial configuration via (\ref{eq: closeness}), the motion ranges of joints $6$ and $7$ of $\bm{\xi}_A$ are much smaller than the corresponding ranges of $\bm{\xi}_N$. Consequently, this could lead to a more time-efficient optimal trajectory. A video of several experiments for comparison can be found in the supplementary material in \url{https://www.acin.tuwien.ac.at/en/360e/}. 
}

Finally, a Monte Carlo simulation is performed to validate the efficiency of the proposed NN in the PTP trajectory optimization. To this end, $10^5$ pairs of initial robot configurations $\mathbf{q}_0$ and target poses $\mathbf{T}_{0,d}^{e}$ are randomly selected from a uniform random distribution in the admissible ranges. Then, 
the proposed NN and the numerical IK are used to determine the target joint configuration and for each target configuration, an optimal trajectory is calculated using (\ref{Eq: discrete}). The statistical results are summarized in Tab. \ref{tab: stat traj}. 
\begin{table}
\caption{Performance of the proposed NN and the numerical IK \cite{lynch2017modern} in the trajectory optimization framework} 
\label{tab: stat traj}
\begin{center}
\scalebox{0.96}{
\begin{tabular}{c c c}
 &  Proposed NN  & Numerical IK \cite{lynch2017modern}\\ 
\hline
Avg. $t_F$ (\SI{}{\second}) & $4.52 \pm 1.93$ &  $5.39 \pm 2.6$\\  
Cost value of (\ref{Eq: discrete a}) & $5.75\pm 2.79$ &  $6.69 \pm 3.29$  \\  
Num. of failed IK & $0$ & $13896$ \\
Num. of failed PTP & $554$ & $1588$\\
Success rate & $99.5\%$ &  $84.5\%$ \\  
\makecell{Avg. comp. time (\SI{}{\milli\second})\\
of (\ref{Eq: discrete})}&  $28.9 \pm 13$ & $30.3 \pm 19$ \\  
\hline
\end{tabular}
}
\end{center}
\end{table}
While the computing times of (\ref{Eq: discrete}) utilizing the target configuration of the proposed algorithm $\mathbf{q}_{t_F,A}$ and the numerical IK $\mathbf{q}_{t_F,N}$ are nearly the same ($\approx$ \SI{30}{\milli\second}), the average optimal trajectory time using the proposed algorithm is slightly better, i.e. \SI{4.52}{\second} compared to \SI{5.39}{\second}. 

Since the solution of the numerical IK depends on the initial guess, $13896$ test cases fail to converge to feasible target configurations. Note that the maximum number of iterations for the numerical IK is $50$. Additionally, after excluding $13896$ failed cases, $1588$ test cases are not valid to plan the PTP trajectory using (\ref{Eq: discrete}). Note that these test cases fail because of violating the iteration limit, i.e., $100$ iterations, which is set in the IPOPT solver. The overall success rate by using the numerical IK is approximately $84.5\%$.
On the other hand, for the proposed algorithm, in all the test cases, a feasible target configuration is found. Only 554 test cases fail during the planning of the PTP trajectory due to the iteration limit of the IPOPT solver. Overall, the proposed NN outperforms the numerical IK by achieving a success rate of $99.5\%$.




\section{Conclusions}
\label{section: conclusion}
In this work, a machine learning-based approach for the inverse kinematics (IK) of kinematically redundant robots is presented, which is suitable for trajectory planning in highly dynamic real-time applications like human-robot object handovers or robotic object catching. 
In this approach, the optimal redundancy parameters are predicted by a neural network (NN) according to the application-specific criteria, closeness to the initial robot configuration and manipulability at the target pose.
Redundancy parameters, i.e. a virtual arm angle and binary variables describing the joint configurations, resolve the non-uniqueness of the analytical IK of redundant robots and allow for a unique mapping between the target pose and the joint configuration.
Since a NN is employed, the proposed framework can be applied to different collaborative robots, e.g., KUKA LBR iiwa 14 R820, Franka Emika Panda, OB7, of which the analytical IK can be parameterized by redundancy parameters.
The NN used in the proposed framework outperforms classical classification algorithms in terms of accuracy 
 and the prediction run time.
A Monte Carlo simulation of $10^5$ random pairs of an initial configuration and a target pose validates the proposed algorithm in the context of point-to-point (PTP) trajectory optimization. 
The proposed method succeeds in $99.5\%$ of the test cases to find a feasible target configuration while achieving a shorter optimal time of the trajectory from the initial to the target pose on average compared to using a numerical IK method at a significantly shorter computing time ($\approx 32$ \SI{}{\micro\second} for the proposed IK compared to $\approx 3$ \SI{}{\milli\second} for the numerical IK). 
Thus, the proposed framework is perfectly suited for real-time applications.

In future works, this machine learning-based framework will be applied to dynamic human-robot handover tasks.
\section*{Appendix}
\label{Appendix 1}
The square of the manipulability (\ref{eq: manipulability-classical}) of the KUKA LBR iiwa 14 R820 \cite{beck2022singlularity} reads as
\begin{equation}
\begin{aligned}
m^2&(\mathbf{q}) = 2d^2_{se}d^2_{ew}\sin^2(q_4) \cdot \\
\bigg[&d^2_{se}\sin^2(q_2)\sin^2(q_4)\cos^2(q_5)\cos^2(q_6)+ \\
&d^2_{ew}\cos^2(q_2)\cos^2(q_3)\sin^2(q_4)\sin^2(q_6)+ \\
&\big(d^2_{se} + 2d_{se}d_{ew}\cos(q_4)-d_{ew}^2\big)\sin^2(q_2)\sin^2(q_6)+\\
&\dfrac{1}{2}\big(d^2_{se}\cos(q_4)+d_{se}d_{ew}\big)\sin^2(q_2)\sin(q_4)\cos(q_5)\sin(2q_6)+\\
&\dfrac{1}{2}\big(d^2_{ew}\cos(q_4)+d_{se}d_{ew}\big)\sin(2q_2)\cos(q_3)\sin(q_4)\sin^2(q_6)\bigg],
\end{aligned}
\label{eq: mani anal}
\end{equation}
where $d_{se} = d_2 + d_3$ and $d_{ew} = d_4 + d_5$.
\section*{Conflict of interest}
The authors have no conflicts of interest to declare.

\bibliographystyle{elsarticle-harv}
\bibliography{ifacconf}           
\end{document}